%% file: acl_latex.tex
\documentclass[11pt]{article}

\usepackage[preprint]{acl}

\usepackage{times}
\usepackage{latexsym}

\usepackage[T1]{fontenc}

\usepackage[utf8]{inputenc}

\usepackage{microtype}

\usepackage{inconsolata}

\usepackage{graphicx}

\usepackage{booktabs}
\usepackage{multirow}
\usepackage{colortbl}
\usepackage{amsmath}
\usepackage{amssymb}
\usepackage{makecell}
\usepackage{soul}
\usepackage{enumitem}
\usepackage[most]{tcolorbox}

\usepackage{pgfplotstable}
\usepackage{subfigure}
\usepackage{pgfplots}
\usepackage{tikz}
\usetikzlibrary{arrows.meta}
\usetikzlibrary{decorations}
\usetikzlibrary{pgfplots.groupplots}
\usepgfplotslibrary{fillbetween}
\usetikzlibrary{shapes.multipart}
\usetikzlibrary{shapes.geometric}

\definecolor{ugreen}{cmyk}{1,0,1,0.498}
\definecolor{lyyblue}{cmyk}{0.8278,0.3333,0,0.2941}
\definecolor{lyygreen}{cmyk}{0.6813,0,0.725,0.3725}
\definecolor{lyyred}{cmyk}{0,0.8855,0.8767,0.1098}
\definecolor{dblue}{cmyk}{1,0.5487,0,0.5569}
\definecolor{royalblue}{HTML}{4169e1}
\definecolor{myred}{HTML}{E33222}
\definecolor{ourblue}{RGB}{123,201,240}
\definecolor{ourgreen}{RGB}{102,164,66}

\newcommand{\method}{SimMIA}
\newcommand{\benchmark}{WikiMIA-25}

\title{Frustratingly Simple Black-Box Membership Inference Attacks\\Against Large Language Models}
\title{Membership Inference on LLMs in the Wild}

\author{Jiatong Yi\thanks{Equal contribution}, Yanyang Li$^*$\\
The Chinese University of Hong Kong \\
\texttt{\{jtyi2, yyli21\}@cse.cuhk.edu.hk}}

\begin{document}
\maketitle
\begin{abstract}
Membership Inference Attacks (MIAs) act as a crucial auditing tool for the opaque training data of Large Language Models (LLMs). However, existing techniques predominantly rely on inaccessible model internals (e.g., logits) or suffer from poor generalization across domains in strict black-box settings where only generated text is available. In this work, we propose \method{}, a robust MIA framework tailored for this text-only regime by leveraging an advanced sampling strategy and scoring mechanism. Furthermore, we present \benchmark{}, a new benchmark curated to evaluate MIA performance on modern proprietary LLMs. Experiments demonstrate that \method{} achieves state-of-the-art results in the black-box setting, rivaling baselines that exploit internal model information\footnote{\url{https://github.com/simmia2026/SimMIA}}.
\end{abstract}

\input{figures/compare}

\section{Introduction}

The unprecedented success of Large Language Models (LLMs) is fueled by pretraining on massive-scale corpora~\citep{qwen3,gpt-5,gemini-2_5}.
However, as state-of-the-art (SOTA) models become increasingly opaque, details regarding their training data are rarely disclosed.
This lack of transparency poses significant challenges for auditing potential copyright infringement, privacy leakage, and data contamination.

A promising avenue to address these concerns is pretraining data detection~\citep{min-k,DBLP:journals/corr/abs-2311-06062}, which typically employs Membership Inference Attacks (MIAs) to determine whether a specific text was part of a model's training set~\citep{DBLP:conf/sp/ShokriSSS17}.
The majority of existing MIA techniques for LLMs are \emph{logit-based}, relying on the assumption that the model exposes per-token probabilities~\citep{min-k-plus-plus,recall}.
While effective for open-weight models, this assumption breaks down for API-accessed LLMs, where typically only the generated text is returned.

To mitigate reliance on model internals, early explorations proposed to estimate logits via linear regression~\citep{petal} or infer membership based on tokenizer distinctiveness~\citep{tong2025membership}.
Despite their utility, these methods often require knowledge of the target model's tokenizer, information that remains proprietary for many closed-source systems.
Therefore, recent research has pivoted toward a strictly \emph{black-box} setting, where the attacker has access only to generated text~\citep{cdd,samia}.
Current black-box methods predominantly rely on approximating likelihoods via N-gram matching.
However, we identify that these approximation methods suffer from two critical limitations: \emph{distribution drift}, where long-form generation diverges from the original distribution, and \emph{signal sparsity}, where exact N-gram matching fails to capture semantic retention, offering unstable estimates across domains.

In this work, we propose \method{}, a novel framework that significantly advances MIA performance in the black-box setting.
Unlike prior works that sample full sequences, \method{} employs a \emph{word-by-word sampling} strategy to eliminate error propagation (mitigating drift) and introduces a \emph{semantic scoring mechanism} that captures soft membership signals beyond exact token matching (mitigating sparsity).
We evaluate \method{} on two standard benchmarks, WikiMIA~\citep{min-k} and MIMIR~\citep{mimir}, across four representative open-source families.
Our results demonstrate that \method{} outperforms existing black-box baselines by a significant margin of 15.7 AUC points on average and achieves parity with, or even surpasses, many logit-based approaches.

To further validate \method{}'s effectiveness on modern proprietary LLMs, we follow WikiMIA-24~\citep{DBLP:conf/aaai/0005W0L0025} to curate \benchmark{}, a dataset derived from 2025 Wikipedia dumps.
Experiments on \benchmark{} with the four latest LLMs show that \method{} establishes a new SOTA in the black-box setting, offering a practical tool for auditing the black-box LLMs of today.

\section{SimMIA}

\subsection{Problem Statement}

Let $M$ be a target LLM trained on a dataset $D$. Given a text $x$, the goal of MIA is to determine whether $x \in D$.
Existing LLM MIAs formalize this as constructing a scoring function over $x$, where a higher score indicates a higher likelihood of membership.
Based on the information exposed by $M$, we distinguish between two threat models:
\begin{itemize}[noitemsep, nolistsep]
    \item \textbf{Gray-box Setting}: In this scenario, the attacker has access to the model's internal tokenization results and even per-token logits. Most existing MIAs operate under this assumption~\citep{min-k,nearest-neighbours,recall,petal}.
    \item \textbf{Black-box Setting}: This setting represents the restricted access typical of commercial APIs. The attacker has no access to model weights, logits, or the tokenization results. The attacker can only observe the generated textual continuation~\citep{cdd,samia}.
\end{itemize}
In this work, we focus on the strictly constrained black-box setting, aiming to perform MIA solely through text-based interaction.




\subsection{Method}

The core idea behind \method{} is that if a text $x$ has appeared in the training data of $M$, then continuations sampled from $M$ given any prefix of $x$ should be more \emph{semantically consistent} with the ground truth suffix than those for unseen examples.
\method{} operationalizes this idea in three stages:
(1) word-by-word sampling,
(2) computation of word-level scores, and
(3) relative aggregation of these scores into a final membership score.
Figure~\ref{fig:compare} illustrates the workflow of \method{} and compares it with the SOTA black-box MIA baseline.

\paragraph{Word-by-Word Sampling}

Previous sampling-based MIA methods~\citep{cdd,samia} typically generate multiple full continuations $x_{\ge i}$ conditioned on a fixed prefix $x_{<i}$ and use them to estimate likelihood-based scores.
However, such approaches suffer from \emph{distribution shift}: as the generation proceeds, later tokens depend on both the true prefix and the model’s self-generated outputs, making the estimated distribution deviate from that of the original text $x$.
\method{} mitigates this issue by sampling only the immediate next word, which avoids self-conditioning and provides an unbiased estimate of the model’s local behavior without increasing the sampling cost.

Specifically, we first tokenize $x$ into a sequence of words $x_1, x_2, \ldots, x_L$ using a standard word tokenizer~\citep{DBLP:books/daglib/0022921}, where $L$ is the number of words.
Starting from the second word, \method{} draws $N$ samples of the next word $\hat{x}_{i}$ from the conditional distribution $M(\cdot \mid x_{<i})$, and computes a word-level score for $x_i$ based on these samples.

\paragraph{Semantic Scoring}

A straightforward method for scoring the target word $x_i$ involves estimating its empirical probability using a set of generated samples $\{\hat{x}^{(j)}_i\}_{j=1}^N$. This formulation aligns with conventional logit-based baselines~\citep{DBLP:conf/csfw/YeomGFJ18,nearest-neighbours,min-k,recall}:
\begin{equation}
s^*(x_i \mid x_{<i}) = \frac{1}{N}\sum_{j=1}^N \mathbb{I}\left[x_i = \hat{x}^{(j)}_i\right],
\label{eqn:empirical}
\end{equation}
where $\mathbb{I}[\cdot]$ is the indicator function.
Laplace smoothing is applied to prevent zero-probability estimates.
We denote this version as \method{}$^*$.

However, this approach relies on a hard indicator, which is inherently sparse due to the high dimensionality of the vocabulary.
This way also fails to utilize the information present in sampled words that do not exactly match $x_i$. Consequently, it suffers from high variance across domains (See Section~\ref{sec:exp}). To overcome this, \method{} introduces a \emph{soft scoring function}.
The key intuition is that if $x$ is part of the training data of $M$, the model should assign high probabilities not only to $x_i$ itself but also to words that are \emph{semantically similar} to it.
To capture this proximity, we define the similarity between two words $a$ and $b$ as:
\begin{equation}
    \mathrm{sim}(a,b) = \frac{\mathrm{cos}\!\left(\mathrm{Enc}(a), \mathrm{Enc}(b)\right) + 1}{2},
    \label{eqn:sim}
\end{equation}
where $\mathrm{Enc}(\cdot)$ denotes the word embedding and $\mathrm{cos}(\cdot)$ computes cosine similarity.
The word-level score is then given by
\begin{equation}
    s(x_i \mid x_{<i}) = \mathbb{E}_{\hat{x}_i \sim M(\cdot \mid x_{<i})}[\mathrm{sim}(x_i, \hat{x}_i)],
    \label{eqn:word-score}
\end{equation}
where the expectation is approximated by the sample mean over $\{\hat{x}_i^{(j)}\}^N_{j=1}$.
This soft formulation leverages both the empirical frequency of $x_i$ and the semantic alignment of non-exact matches, thereby enhancing robustness across domains.

\begin{figure}[t]
    \includegraphics[width=\linewidth]{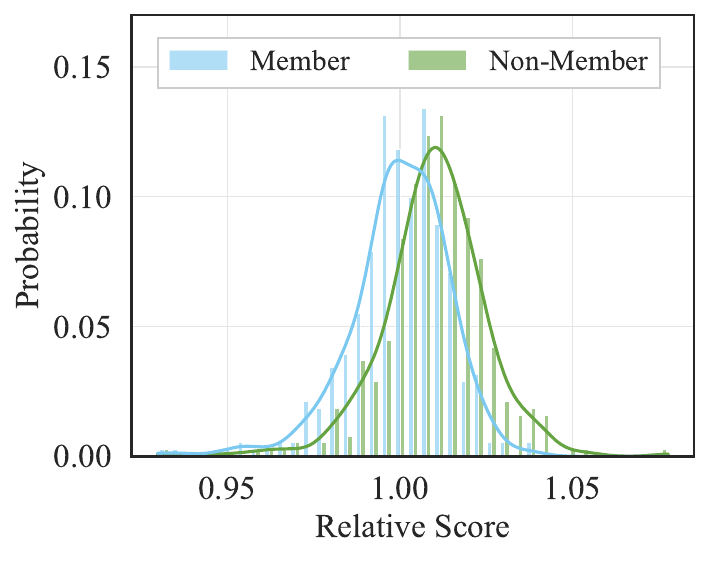}
    \caption{Relative score distributions for members and non-members. The relative score is the ratio of the perturbed score to the unperturbed score in Eq.~\ref{eqn:word-score}. Consistent with \citet{recall}, members show lower relative scores compared to non-members.}
    \label{fig:dist}
    \vspace{-10pt}
\end{figure}

\paragraph{Relative Aggregation}

Finally, we aggregate word-level scores into an overall membership score for $x$.
\citet{recall} demonstrated that the likelihood of member data is significantly lower than that of non-member data when perturbed by non-members.
We observe a similar phenomenon in the relative score distribution of $s(x_i \mid x_{<i})$ in Eq.~\ref{eqn:word-score}, as illustrated in Figure~\ref{fig:dist}.
Drawing on this insight, we adopt a formulation similar to \citet{recall} and define the final membership score as:
\begin{equation}
    \mathrm{SimMIA}(x) = -\frac{1}{L}\sum_{i=1}^L 
    \frac{s(x_i \mid P \oplus x_{<i})}{s(x_i \mid x_{<i})},
    \label{eqn:relative}
\end{equation}
where $\oplus$ is the string concatenation and $P = p_1 \oplus p_2 \oplus \cdots \oplus p_T$ denotes the concatenation of $T$ non-member samples used to prefix the context.

\input{tables/wikimia}
\input{tables/mimir}

\section{WikiMIA-25}

While numerous MIA benchmarks have been proposed, they are often limited to testing specific LLMs.
For example, WikiMIA~\citep{min-k} applies only to models released between 2017 and 2023, while MIMIR~\citep{mimir} is applicable only to models trained on The Pile~\citep{pile}.
Given the rapid release cadence of LLMs, benchmarks require frequent updates to remain relevant.
For instance, WikiMIA-24~\citep{DBLP:conf/aaai/0005W0L0025} extends the validity period of WikiMIA to 2024.

Moreover, evaluating proprietary LLMs presents a unique challenge: unlike open-source models, which remain accessible indefinitely, proprietary models are hosted on API endpoints subject to vendor deprecation.
This creates a gap: as vendors retire older model versions that fit a benchmark's timeframe, they release newer versions trained on data that post-dates the benchmark.
This temporal misalignment renders the benchmark ineffective.
For instance, while WikiMIA-24 updates its data to March 2024, the oldest Gemini version available by the time of conducting experiments, Gemini-2.0-Flash, possesses a knowledge cutoff of June 2024.
Since the model has likely seen the benchmark's data during training, WikiMIA-24 cannot reliably evaluate Gemini.

To evaluate MIA on the latest LLMs, we introduce \benchmark{}.
Similar to WikiMIA and WikiMIA-24, we construct datasets from recent Wikipedia dumps.
We designate events added from January 2015 to December 2016 as member data, and events from March 2025 to September 2025 as non-member data.
We truncate examples to $L$ whitespace-delimited tokens, creating buckets where $L\in\{32,64,128\}$.
The final benchmark comprises 759, 875, and 529 pairs of member and non-member examples for the respective lengths.



\section{Experiments}
\label{sec:exp}

\subsection{Setup}

\paragraph{Benchmarks}
We evaluate our method on WikiMIA~\citep{min-k} and MIMIR~\citep{mimir}, alongside our \benchmark{}.
For WikiMIA, we utilize subsets with sequence lengths of 32, 64, and 128.
For MIMIR, we report results on the 7-gram version.
To ensure cost efficiency on \benchmark{}, we sample a balanced subset of 120 members and 120 non-members from the length-32 split.
For all benchmarks, we reserve 10 non-members to construct the non-member prefix $P$ and evaluate on the remaining examples.

Consistent with prior work~\citep{min-k,min-k-plus-plus}, we report the Area Under the ROC Curve (AUC) as our primary metric, along with the True Positive Rate at 5\% False Positive Rate (TPR@5\%FPR).

\paragraph{Baselines}
We compare \method{} against ten MIA baselines across gray-box and black-box settings.
In the \emph{gray-box} setting, we include:
\textbf{Loss}, \textbf{Zlib}, and \textbf{Lowercase}~\citep{losszlib}, which utilize the raw input loss, the input's Zlib entropy, and the lower-cased input loss, respectively;
\textbf{Reference}~\citep{smallerref}, which calibrates loss using a smaller model;
\textbf{Neighbors}~\citep{nearest-neighbours}, which compares input loss to that of similar tokens;
\textbf{Min-K\%}~\citep{min-k} and \textbf{Min-K\%++}~\citep{min-k-plus-plus}, which focus on low-probability tokens;
\textbf{ReCaLL}~\citep{recall}, which measures relative conditional likelihood;
and \textbf{PETAL}~\citep{petal}, which uses regression based on token-label similarity.
In the \emph{black-box} setting, we evaluate \textbf{SaMIA}~\citep{samia}, which approximates likelihood via ROUGE-N scores over sampled continuations.

\input{tables/wikimia-25}

\paragraph{Models}

For WikiMIA, we test \texttt{OPT-6.7B}~\citep{opt}, \texttt{Pythia-6.9B}~\citep{pythia}, \texttt{LLaMA-7B}~\citep{llama}, and \texttt{GPT-NeoX-20B}~\citep{gpt-neox}.
For MIMIR, we follow~\citet{mimir} and focus on the \texttt{Pythia-dedup} family (160M, 1.4B, 2.8B, and 6.9B).
For \benchmark{}, we extend our evaluation beyond \texttt{Pythia-6.9B} to newer models, including \texttt{Qwen3-8B-Base}~\citep{qwen3}, \texttt{Claude-4.5-Haiku} ~\citep{claude-haiku-4_5}, \texttt{Gemini-2.5-Flash}~\citep{gemini-2_5}, and \texttt{GPT-5-Chat}~\citep{gpt-5}.
\paragraph{Implementation Details}
We reproduce all baselines using their official hyperparameter settings.
For \method{}, we set the sample size $N=100$, except for \benchmark{} where we reduce $N$ to 10 to mitigate API costs.
The number of shots $T$ is set to 7 for WikiMIA and \benchmark{}, and 10 for MIMIR.
We use \texttt{all-MiniLM-L6-v2}\footnote{\url{https://huggingface.co/sentence-transformers/all-MiniLM-L6-v2}} for embedding~\citep{petal}.
A specific exception is applied to the DM Mathematics split of MIMIR, where we employ Exact Match for numerical tokens.
In mathematical reasoning, numerical proximity does not imply semantic equivalence (e.g., distinct result values indicate incorrect reasoning regardless of magnitude); therefore, we strictly enforce exact matching for numbers.
All experiments are conducted on 8 NVIDIA RTX 3090.

\input{tables/ablation_study}

\subsection{Main Results}
\noindent\textbf{\method{} achieves SOTA black-box MIA performance and rivals gray-box counterparts.}
Tables~\ref{tab:wikimia} and \ref{tab:mimir} demonstrate that \method{} substantially outperforms existing black-box baselines, improving AUC by 16.6 and 14.9 on WikiMIA and MIMIR, respectively.
Remarkably, \method{} bridges the gap with SOTA gray-box methods; it trails the best gray-box performance by only 6.1 and 3.4 AUC points on average.
This margin is comparable to the difference between the first- and second-best gray-box methods (a gap of 10.2 and 0.7 AUC), suggesting that \method{} remains highly competitive despite lacking access to model internals.
It even surpasses the best gray-box method on specific LLMs, such as OPT-6.7B on WikiMIA.

\input{figures/num_shots_samples}

\noindent\textbf{\method{} generalizes effectively to the latest LLMs.}
Table~\ref{tab:wikimia25} highlights that \method{} delivers superior black-box MIA results on both legacy models (e.g., \texttt{Pythia-6.9B}) and recent releases (e.g., \texttt{Qwen3-8B-Base}) on \benchmark{}.
Crucially, \method{} extends to proprietary LLMs where gray-box attacks are inapplicable.
In this strict black-box setting, \method{} significantly surpasses SaMIA, the current SOTA black-box MIA, achieving gains of +21.7 in AUC and +25.8 in TPR@5\%FPR.
Detailed evaluations of these proprietary models are provided in Appendix~\ref{app:template}.

\noindent\textbf{The soft scoring function ensures robustness across benchmarks.}
Directly estimating label probabilities (\method{}$^*$) yields peak performance on WikiMIA-style datasets (WikiMIA and \benchmark{}), outperforming SaMIA, the SOTA black-box baseline, by +19.2 AUC.
However, we find that employing our soft scoring function (\method{}) offers superior generalization and stability across diverse benchmarks.
As shown in Tables~\ref{tab:wikimia}, \ref{tab:mimir}, and \ref{tab:wikimia25}, \method{} consistently outperforms SaMIA by a significant margin in all settings.
Crucially, while \method{}$^*$ surpasses SaMIA by only +7.6 AUC on MIMIR, \method{} remains robust, outperforming SaMIA by +14.9 AUC on MIMIR while maintaining high efficacy on WikiMIA-style datasets (+15.4 AUC).

\input{tables/embedding_models}

\subsection{Analysis}
In this section, we provide a detailed analysis of \method{}.
Unless otherwise stated, we report AUC results on WikiMIA using \texttt{Pythia-6.9B}.

\paragraph{Ablation Study}
We conduct an ablation study to quantify the contribution of each core component in \method{}. Specifically, we examine two variants: (1) \textit{w/o Word-by-Word Sampling}, where we replace our granular sampling strategy with SaMIA-style generation (producing $N$ continuations conditioned on the first half of the input), and (2) \textit{w/o Relative Aggregation}, where we utilize Eq.~\ref{eqn:word-score} directly without normalizing against the non-member conditioned score.
As detailed in Table~\ref{tab:ablation}, eliminating either component results in a significant performance drop.
This underscores the necessity of our designs.


\input{tables/robustness}

\paragraph{Impact of Prefix Length}
Since black-box MIAs often rely on continuations generated from a prefix, we investigate how the length of this prefix affects performance.
Intuitively, performance is expected to be inversely proportional to the prefix ratio: as the prefix length increases (from 10\% to 90\%), the remaining target segment becomes shorter, providing less signal for the MIA to exploit.
We compare \method{} (which defaults to conditioning on the first word) against SaMIA (which defaults to the first 50\%).
As shown in the left panel of Figure~\ref{fig:combined_numshots}, \method{} aligns with the expected behavior: performance peaks at a 10\% prefix ratio and declines monotonically as the prefix grows.
In contrast, SaMIA shows no clear gains from shorter prefixes: it attains its best performance at a 20\% prefix ratio, then fluctuates across intermediate ratios before dropping at 70\%.
This suggests that SaMIA fails to exploit the rich signal available in the early portions of the text, whereas \method{} remains robust even with minimal prefix conditioning.

\paragraph{Impact of Sample Size}
We examine the trade-off between computational budget (sample size $N$) and inference accuracy.
The middle panel of Figure~\ref{fig:combined_numshots} demonstrates that \method{} consistently outperforms SaMIA across all sample sizes.
Notably, \method{} exhibits better scalability: it continues to accrue performance gains up to $N=30$. Conversely, SaMIA saturates rapidly, failing to yield improvements beyond $N=10$.

\paragraph{Sensitivity to Hyperparameters}
\method{} introduces two design choices: the number of non-member shots ($T$) and the embedding model.
In the right panel of Figure~\ref{fig:combined_numshots}, we observe that inference performance improves as $T$ increases, stabilizing around $T=7$. This suggests that while non-member shots are crucial, a small, computationally efficient set is sufficient.
Regarding the embedding model, Table~\ref{tab:different-embedding-model} reports performance across a spectrum of architectures, ranging from static embeddings (Word2Vec~\citep{word2vec}, fastText~\citep{fasttext}) to state-of-the-art dense retrievers (e.g., \texttt{bge-large-en-v1.5}~\citep{DBLP:journals/corr/abs-2309-07597},
\texttt{UAE-Large-V1}~\citep{DBLP:conf/acl/LiL24}, and \texttt{mxbai-embed-large-v1}~\citep{emb2024mxbai}).
Notably, although dense retrievers yield superior performance, \method{} maintains high stability regardless of the underlying embedding model choice.

\paragraph{Stability and Variance}
We analyze the performance variance stemming from randomness in non-member shot selection (fixed vs. random) and sampling seeds.
Table~\ref{tab:robustness} indicates that while \method{} is robust across different sampling seeds, it exhibits sensitivity to the composition of non-member prefixes.
Specifically, randomizing prefixes results in a higher standard deviation ($\sigma\approx$ 2.90) compared to using a fixed set ($\sigma\approx$ 0.86).
Crucially, however, despite this increased fluctuation, the average performance of \method{} with random prefixes still significantly outperforms SaMIA (72.77 vs. 60.38), demonstrating the method's effectiveness even under suboptimal settings.

\input{tables/prefix_set}

\paragraph{Impact of Prefix Similarity}
We hypothesize that the variance in random selection arises from the varying semantic similarity between the prefix and the test instance~\citep{recall}.
To verify this, we retrieve non-members for each test instance using TF-IDF~\citep{TF-IDF} and partition them into three tiers (\textit{most}, \textit{moderately}, and \textit{least} similar) for comparison against \textit{random} and \textit{fixed} baselines.
Table~\ref{tab:prefix-selection} confirms that performance improves as similarity increases (8.7 AUC on average).
Notably, however, the \textit{fixed} prefix strategy yields performance comparable to the \textit{moderately similar} and \textit{random} settings.
This implies that dynamic retrieval is not strictly necessary; a simple, fixed set of non-member shots is sufficient to capture the primary benefits of \method{}.

\paragraph{Progressive Disclosure}
\method{} has demonstrated its effectiveness in a strict black-box setting, where only generated text is available to the attacker.
However, real-world API access exists on a spectrum; some services may disclose metadata such as tokenization results or token probabilities.
We therefore ask: can \method{} adapt to exploit this additional information for improved detection?

We evaluate \method{} using \texttt{GPT-5-Chat} and \texttt{GPT-4.1-mini} on \benchmark{}. The former provides tokenization results for the input and output, while the latter provides both tokenization results and the log probabilities of generated tokens.
We adapt \method{} as follows:
(1) When tokenization is visible, we shift from word-by-word sampling to token-by-token sampling.
(2) When log probabilities are available, we directly weight $\mathrm{sim}\left(\cdot\right)$ in Eq.~\ref{eqn:word-score} by the ground-truth model probabilities (normalized across unique next tokens).

As shown in Table~\ref{tab:expose}, \method{} yields superior performance when these additional signals are integrated.
This underscores the versatility of \method{}: it remains robust under strict black-box constraints but is flexible enough to leverage richer information from progressively disclosed metadata.

\input{tables/expose}

\section{Related Work}

Membership inference attacks (MIA) determine whether a data point was included in a model’s training set~\citep{DBLP:conf/sp/ShokriSSS17}. 
While a large body of work on LLMs relies on model logits, either directly~\citep{nearest-neighbours,min-k,min-k-plus-plus,dc-pdd} or via relative log-likelihoods~\citep{recall,con-recall}, these methods are inapplicable to proprietary LLMs where APIs typically restrict logit access.

To address this, recent studies have explored more constrained gray-box settings. 
\citet{petal} approximate target logits using an open-source proxy, while \citet{tong2025membership} exploit tokenizer behaviors. 
However, these methods generally presume access to the target's tokenized outputs, which are often unavailable.

Consequently, research has shifted toward black-box settings where only generated text is observable.
Existing approaches mostly target specialized scenarios. 
Quiz-based methods~\citep{dcq,decop} use manual templates to generate QA pairs for detection; 
training-based methods~\citep{veilprobe} train detectors on known members and non-members; 
and set-based strategies~\citep{blackbox-dataset-inference,provenance-mia} infer membership at the dataset level.

Closest to our work are sampling-based methods~\citep{samia,cdd}. 
Yet, these general-purpose approaches heavily rely on surface-form statistics (e.g., N-gram matches), making them brittle across domains. 
\method{} advances this line of work by improving both the sampling strategy and the scoring function.


\section{Conclusion}
In this work, we propose \method{}, a robust framework for MIAs in the strict black-box setting.
We also introduce \benchmark{}, a curated MIA benchmark designed to evaluate modern proprietary LLMs.
Experimental results demonstrate that \method{} achieves SOTA performance, effectively rivaling gray-box counterparts even when model internals are completely inaccessible.

\section*{Limitations}
Despite its effectiveness, \method{} shares a fundamental limitation with other sampling-based black-box MIAs on LLMs: the reliance on sampling to estimate membership signals.
Unlike gray-box counterparts that can derive metrics from a single forward pass, general-purpose black-box methods necessitate repeated querying to approximate model behavior.
This requirement inevitably incurs higher computational overhead, latency, and cost.
Consequently, a critical direction for future work is to enhance the sample efficiency of the entire class of sampling-based black-box MIAs, thereby reducing the budget required for reliable detection.


\bibliography{custom}

\appendix

\section{Prompt Template for Proprietary LLMs in \benchmark{}}
\label{app:template}

Modern proprietary LLMs typically function as instruction-following assistants rather than base language models; as a result, they do not perform text continuation by default.
Since \method{} relies on sampling potential next tokens for a given prefix, explicitly prompting these models to mimic text completion behavior is essential.
Below, we present the specific system instruction used to constrain the models to generate a single-token completion for a \colorbox{blue!10}{prefix}.


\patchcmd{\quote}{\rightmargin}{\leftmargin 15pt \rightmargin}{}{}
\begin{quote}
\small 
\begin{tcolorbox}[breakable, colback=white, colbacktitle=blue!5!white, colframe=black, boxrule=1pt, title={\textcolor{black}{\textbf{Prompt}}}]
\textbf{System:} Return ONLY the single next token from the text. It can be punctuation or one whole word. No spaces, no quotes, no extra text.\\
\textbf{User:} Text so far: \colorbox{blue!10}{[PREFIX]}
\end{tcolorbox}
\end{quote}
\patchcmd{\quote}{\rightmargin}{\leftmargin 26pt \rightmargin}{}{}

\section{Proprietary LLM Version}

\input{tables/close-source-model-versions}

We document the exact model identifiers (version) to support reproducibility. While we pin to fixed snapshots when available, some models only offer aliases; in such cases, results may evolve as the provider updates the underlying system.





\input{tables/prefix_domain}

\section{Impact of Prefix Domain}
While \method{} show promise, we explore whether utilizing non-members from disparate domains yields similar improvements~\citep{recall}.
As shown in Table~\ref{tab:different-prefix-domain}, employing prefixes from out-of-domain sources leads to performance degradation.
This suggests that to maximize the efficacy of \method{}, it is advisable to collect in-domain non-members whenever possible.

\end{document}

%% file: figures/compare.tex
\begin{figure*}[t!]
    \newdimen\base
    \base=0.7cm

    \tikzstyle{textnode} = [rectangle,font=\scriptsize,draw=black,inner sep=0pt,outer sep=0pt,minimum width=2\base,minimum height=\base,rounded corners=2pt]

    \hspace*{\fill}
    \subfigure[SaMIA]
    {
        \centering
        \includegraphics[height=3cm]{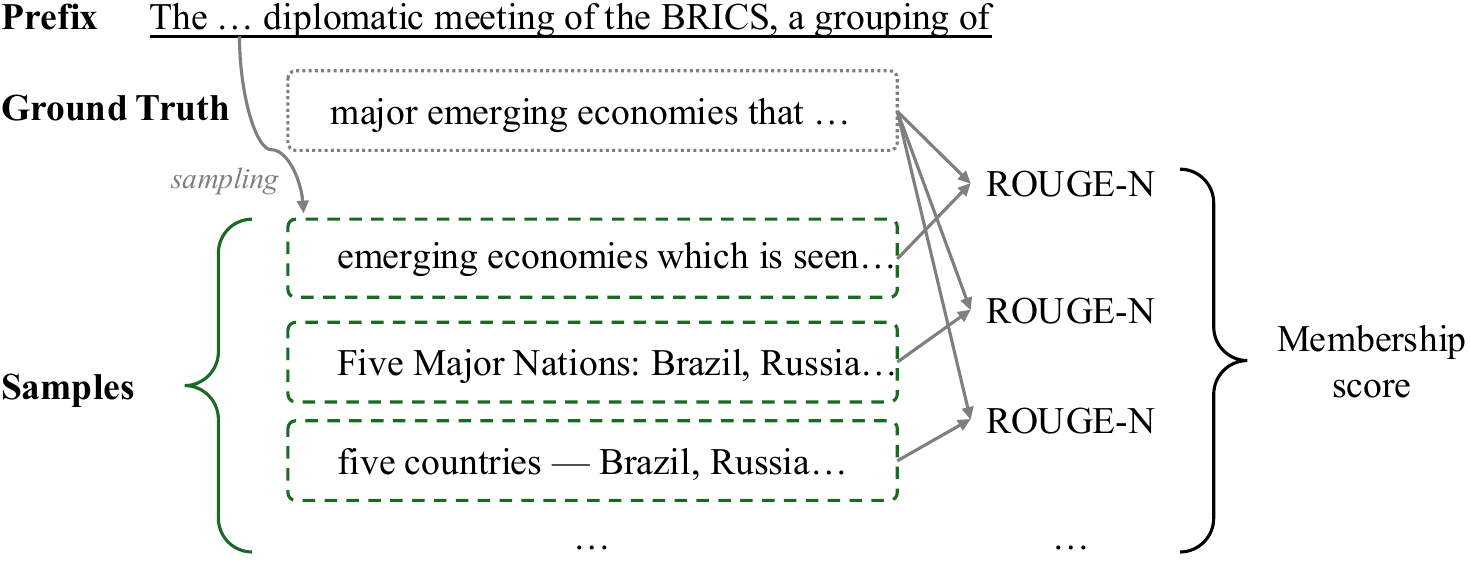}
        \label{fig:samia}
    }
    \hfill
    \rule{0.5pt}{3cm}
    \hfill
    \subfigure[\method{}]
    {
        \centering
        \includegraphics[height=3cm]{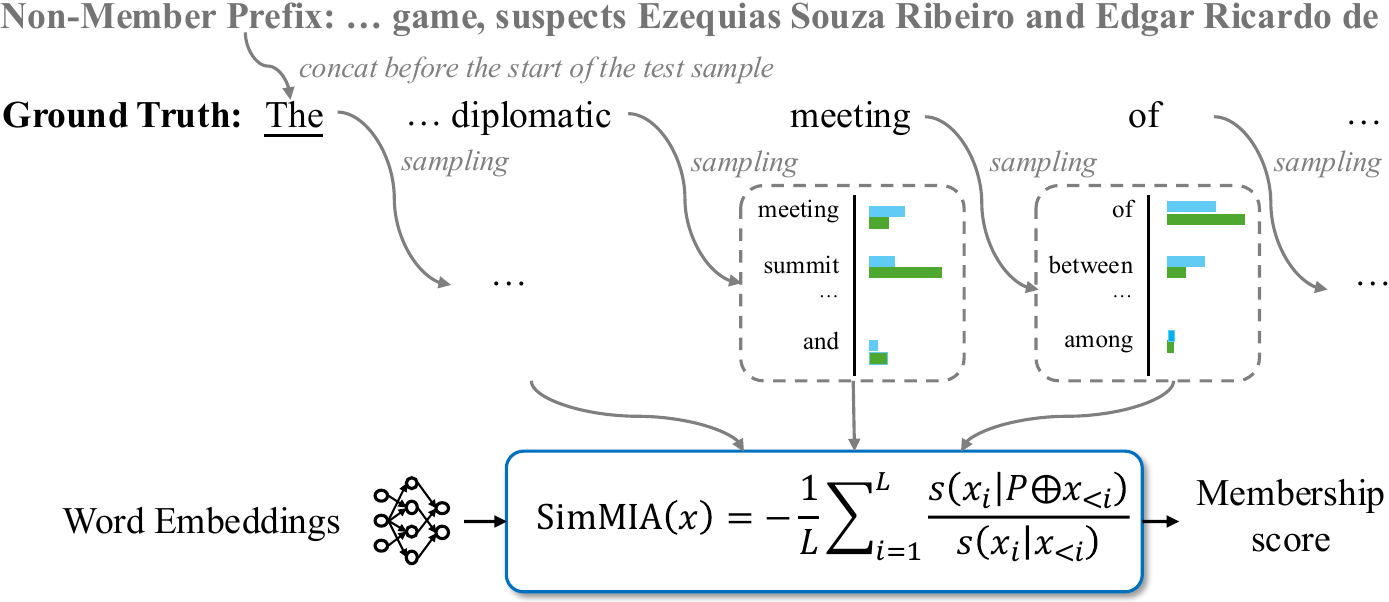}
        \label{fig:simmia}
    }
    \hspace*{\fill}
    \caption{Comparison between SaMIA~\citep{samia}, a representative black-box MIA baseline, and our proposed \method{}. 
Given a text $x$: 
(a) \textbf{SaMIA} samples multiple candidate continuations conditioned on the \ul{underlined} prefix, computes ROUGE-N scores~\citep{lin-2004-rouge} between each candidate and $x$, and aggregates these scores to determine membership; 
(b) \textbf{\method{}} instead performs word-by-word sampling for every prefix, computes word-level scores based on sampled and the corresponding words from $x$ using word embeddings, and aggregates these scores to derive the final membership prediction.
{\protect\tikz \protect\draw[color=ourblue, fill=ourblue] plot[mark=square*, mark options={scale=1.4}] (0,0);} denotes the empirical frequency of the corresponding word, while
{\protect\tikz \protect\draw[color=ourgreen, fill=ourgreen] plot[mark=square*, mark options={scale=1.4}] (0,0);} is the empirical frequency if additionally conditioning on non-member data.}
    \label{fig:compare}
    \vspace{-10pt}
\end{figure*}

%% file: tables/wikimia.tex
\begin{table*}[t!]
\centering
\resizebox{\linewidth}{!}{
\setlength{\tabcolsep}{5pt}
\begin{tabular}[t]{llcccccccccc}
\toprule
\makecell[c]{\multirow{2}{*}{\textbf{Len}}} & \makecell[c]{\multirow{2}{*}{\textbf{Method}}} & \multicolumn{2}{c}{\textbf{OPT-6.7B}} & \multicolumn{2}{c}{\textbf{Pythia-6.9B}} & \multicolumn{2}{c}{\textbf{LLaMA-13B}} & \multicolumn{2}{c}{\textbf{GPT-NeoX-20B}} & \multicolumn{2}{c}{\textbf{Average}} \\
\cmidrule(lr){3-4}\cmidrule(lr){5-6}\cmidrule(lr){7-8}\cmidrule(lr){9-10}\cmidrule(lr){11-12}
&  & AUC & TPR@5\%FPR & AUC & TPR@5\%FPR & AUC & TPR@5\%FPR & AUC & TPR@5\%FPR & AUC & TPR@5\%FPR \\
\midrule
\multirow{14}{*}{32} & \multicolumn{11}{c}{\cellcolor{gray!30}\texttt{Gray-Box}} \\
& Loss & 60.3 & 10.5 & 63.5 & 13.7 & 67.4 & 13.7 & 68.7 & 19.0 & 65.0 & 14.2\\
& Reference & 64.0 & 7.9 & 63.8 & 6.3 & 58.2 & 4.7 & 67.9 & 16.6 & 63.5 & 8.9\\
& Lowercase & 57.7 & 8.9 & 58.9 & 11.8 & 65.3 & 13.4 & 61.6 & 14.5 & 60.9 & 12.2 \\
& Zlib & 61.3 & 12.6 & 64.1 & 16.8 & 67.7 & 11.8 & 69.0 & 20.8 & 65.5 & 15.5\\
& Neighbor & 64.5 & 12.1 & 65.7 & \underline{17.1} & 66.2 & 11.8 & 70.4 & 22.9 & 66.7 & 16.0\\
& Min-K\% & 61.1 & \underline{16.3} & 69.5 & 13.2 & 66.5 & 18.4 & 71.5 & \underline{28.2} & 67.2 & \underline{19.0} \\
& Min-K\%++ & \underline{65.1} & 8.9 & \underline{70.0} & 13.9 & \underline{84.5} & \underline{32.4} & \underline{74.2} & 19.0 & \underline{73.5} & 18.6\\
& ReCaLL & \textbf{77.6} & \textbf{25.5} & \textbf{86.1} & \textbf{28.9} & \textbf{89.1} & \textbf{41.8} & \textbf{87.9} & \textbf{34.7} & \textbf{85.2} & \textbf{32.7}\\
& PETAL & 61.4 & 10.5 & 62.8 & 15.3 & 63.2 & 11.1 & 67.6 & 21.6 & 63.8 & 14.6 \\
\cmidrule{2-12}
& \multicolumn{11}{c}{\cellcolor{gray!30}\texttt{Black-Box}} \\
& SaMIA & 55.4 & 5.3 & 55.2 & 8.7 & \underline{59.4} & 10.0 & 60.1 & 10.8 & 57.5 & 8.7\\
& \method{}$^*$ & \textbf{84.7} & \textbf{18.9} & \textbf{81.7} & \textbf{26.3} & 59.3 & \underline{12.6} & \textbf{84.4} & \textbf{33.4} &\textbf{77.5} & \textbf{22.8}\\
& \method{} & \underline{78.3} & \underline{14.7} & \underline{70.1} & \underline{13.4} & \textbf{62.8} & \textbf{12.9} & \underline{75.5} & \underline{20.3} & \underline{71.7} & \underline{15.3} \\
\midrule
\multirow{14}{*}{64} & \multicolumn{11}{c}{\cellcolor{gray!30}\texttt{Gray-Box}} \\
& Loss & 56.5 & 13.9 & 60.0 & 13.1 & 63.4 & 12.0 & 65.8 & 12.0 & 61.4 & 12.8\\
& Reference & 62.5 & 5.2 & 63.1 & 12.4 & 63.8 & 4.0 & 66.2 & 15.9 & 63.9 & 9.4 \\
& Lowercase & 55.6 & 11.2 & 58.7 & 8.4 & 61.8 & 8.4 & 60.4 & 12.4 & 59.1 & 10.1\\
& Zlib & 59.2 & 12.0 & 62.0 & 15.9 & 65.2 & 12.7 & 67.5 & 17.1 & 63.5 & 14.4\\
& Neighbor & 60.7 & 12.0 & 63.6 & 9.6 & 64.3 & 9.6 & 67.9 & 12.4 & 64.1 & 10.9\\
& Min-K\% & 59.8 & 16.7 & 64.0 & 20.3 & 65.9 & 18.3 & 71.7 & 18.7 & 65.4 & 18.5\\
& Min-K\%++ & \underline{64.9} & \underline{17.1} & \underline{71.7} & \underline{21.9} & \underline{84.6} & \underline{29.5} & \underline{77.2} & \underline{25.5} & \underline{74.6} & \underline{23.5}\\
& ReCaLL & \textbf{76.3} & \textbf{23.9} & \textbf{89.8} & \textbf{49.8} & \textbf{89.5} & \textbf{48.6} & \textbf{87.1} & \textbf{33.5} & \textbf{85.7} & \textbf{39.0}\\
& PETAL & 59.8 & 12.7 & 60.7 & 15.5 & 62.0 & 10.4 & 65.8 & 15.1 & 62.1 & 13.4\\
\cmidrule{2-12}
& \multicolumn{11}{c}{\cellcolor{gray!30}\texttt{Black-Box}} \\
& SaMIA & 65.1 & 10.4 & 61.2 & 8.8 & \underline{63.1} & 10.8 & 67.1 & \underline{11.6} & 64.1 & 10.4 \\
& \method{}$^*$ & \textbf{86.0} & \textbf{25.1} & \textbf{87.5}& \textbf{37.5} & 61.0 & \underline{11.2} & \textbf{86.0} & \textbf{30.3} & \textbf{80.1} & \textbf{26.0}\\
& \method{}& \underline{78.8} & \underline{19.5} & \underline{75.4} & \underline{19.1} & \textbf{66.7} & \textbf{11.6} & \underline{71.5} & 11.2 & \underline{73.1} & \underline{15.4} \\
\midrule
\multirow{14}{*}{128} & \multicolumn{11}{c}{\cellcolor{gray!30}\texttt{Gray-Box}} \\
& Loss & 62.8 & 13.5 & 65.2 & 15.4 & 68.9 & 21.2 & 71.5 & 24.0 & 67.1 & 18.5\\
& Reference & 64.3 & 12.5 & 63.8 & 15.4 & 63.8 & 13.5 & 69.7 & 23.1 & 65.4 & 16.1\\
& Lowercase & 61.2 & 6.7 & 64.8 & 3.8 & 67.6 & 20.2 & 68.0 & 8.7 & 65.4 & 9.9\\
& Zlib & 63.6 & 16.3 & 66.9 & 18.3 & 70.2 & 20.2 & 72.3 & 25.0 & 68.3 & 20.0\\
& Neighbor & 64.0 & 16.3 & 69.3 & 13.5 & 70.3 & 22.1 & 73.6 & 22.1 & 69.3 & 18.5\\
& Min-K\% & 67.6 & 15.4 & 69.5 & \underline{20.2} & 73.3 & 19.2 & 75.9 & 24.0 & 71.6 & 19.7 \\
& Min-K\%++ & \textbf{71.4} & \textbf{25.0} & \underline{72.2} & 19.2 & \underline{86.4} & \underline{38.5} & \underline{76.5} & 28.9 & \underline{76.6} & \underline{27.9}\\
& ReCaLL & \underline{71.0} & \underline{17.3} & \textbf{87.4} & \textbf{41.3} & \textbf{91.3} & \textbf{48.1} & \textbf{88.0} & \textbf{39.4} & \textbf{84.4} & \textbf{36.5}\\
& PETAL & 64.8 & 12.5 & 65.2 & 19.2 & 64.3 & 13.5 & 71.5 & \underline{34.6} & 66.5 & 20.0\\
\cmidrule{2-12}
& \multicolumn{11}{c}{\cellcolor{gray!30}\texttt{Black-Box}} \\
& SaMIA & 67.2 & \underline{11.5} & 62.4 & 7.7 & \textbf{65.2} & \underline{7.7} & 67.5 & 16.3 & 65.6 & 10.8\\
& \method{}$^*$ & \textbf{83.7} & 5.8 & \textbf{89.4} & \textbf{45.2} & 61.3 & \underline{7.7} & \textbf{83.2} & \textbf{28.8} & \textbf{79.4} & \textbf{21.9}\\
& \method{} & \underline{78.8} & \textbf{19.5} & \underline{73.2} & \underline{17.3} & \underline{62.7} & \textbf{14.4} & \underline{74.0} & \underline{17.3} & \underline{72.2} & \underline{17.1} \\
\bottomrule
\end{tabular}
}
\caption{Results on WikiMIA benchmark. \textbf{Bolded} and \underline{underlined} numbers are the best and second-best results within each column for the given length and setting. $^*$ means that \method{} uses Eq.~\ref{eqn:empirical} rather than Eq.~\ref{eqn:word-score} for scoring.}
\label{tab:wikimia}
\vspace{-10pt}
\end{table*}

%% file: tables/mimir.tex
\begin{table*}[t!]
\centering
\resizebox{\linewidth}{!}{
\setlength{\tabcolsep}{5pt}
\begin{tabular}[t]{lcccccccccccccccc}
\toprule
\makecell[c]{\multirow{2}{*}{\textbf{Method}}} & \multicolumn{4}{c}{\textbf{Wikipedia}} & \multicolumn{4}{c}{\textbf{Github}} & \multicolumn{4}{c}{\textbf{Pile CC}} & \multicolumn{4}{c}{\textbf{PubMed Central}} \\
\cmidrule(lr){2-5}\cmidrule(lr){6-9}\cmidrule(lr){10-13}\cmidrule(lr){14-17}
& 160M & 1.4B & 2.8B & 6.9B & 160M & 1.4B & 2.8B & 6.9B & 160M & 1.4B & 2.8B & 6.9B & 160M & 1.4B & 2.8B & 6.9B \\
\midrule
\rowcolor{gray!30}
\multicolumn{17}{c}{\texttt{Gray-Box}}\\
Loss & \textbf{62.3} & \textbf{64.9} & \underline{66.2} & \underline{66.4} & 83.6 & 86.4 & 88.4 & 87.9 & \textbf{53.7}& 55.2 & \underline{54.9} & 57.0 & \underline{79.1} & \textbf{78.0} & \textbf{77.8} & \textbf{77.6}\\
Reference & 51.7 & 62.9 & \textbf{67.6} & \textbf{67.8} & 67.7 & 73.3 & 73.7 & 65.7 & 52.0 & \textbf{58.6} & \textbf{58.2} & \textbf{63.7} & 68.0 & 66.8 & 63.0 & 60.5\\
Lowercase & 58.6 & 62.1 & 63.5 & 64.1 & 76.5 & 81.2 & 83.5 & 83.3 & \underline{53.6} & 54.3 & 54.2 & 55.7 & 75.1 & 75.2 & 75.4 & 75.2\\
Zlib & 56.5 & 60.9 & 63.0 & 63.0 & \textbf{87.9} & \textbf{89.8} & \textbf{91.2} & \textbf{90.9} & 51.9 & 53.9 & 53.7 & 55.6 & 77.8 & 76.9 & \underline{76.7} & \underline{76.7}\\
Neighbor & 57.9 & 60.9 & 61.8 & 61.7 & \underline{86.6} & \underline{88.8} & \underline{90.5} & \underline{89.4} & 52.7 & 55.2 & 54.8 & 56.8 & 76.8 & 74.4 & 74.0 & 74.4\\
Min-K\% & 59.9 & \underline{63.8} & 65.6 & 65.8 & 82.9 & 86.5 & 88.4 & 88.3 & 52.9 & \underline{55.3} & 54.7 & 56.9 & 77.3 & \underline{77.4} & \textbf{77.8} & \textbf{77.6}\\
Min-K\%++ & 55.2 & 62.1 & 64.2 & 63.8 & 72.5 & 81.6 & 85.0 & 84.9 & 51.6 & 54.9 & 54.1 & 56.3 & 62.3 & 63.0 & 66.0 & 67.5\\
ReCaLL & 55.9 & 63.4 & 65.1 & 65.2 & 83.0 & 86.0 & 87.9 & 87.4 & 51.3 & 54.7 & 53.7 & \underline{57.4} & \textbf{79.5} & 75.2 & 76.0 & 73.7 \\
PETAL & \underline{60.1} & 62.4 & 63.7 & 63.6 & 68.2 & 70.7 & 72.1 & 71.3 & 52.8 & 53.9 & 54.0 & 55.4 & 72.1 & 72.7 & 73.3 & 73.4\\
\midrule
\rowcolor{gray!30}
\multicolumn{17}{c}{\texttt{Black-Box}}\\
SaMIA & 49.1 & 51.3 & 51.9 & 52.1 & 53.6 & 60.3 & 61.1 & 62.7 & 50.1 & 48.9 & 48.3 & 49.9 & \underline{53.1} & 51.2 & 54.8 & 52.6\\
\method{}$^*$ & \underline{57.3} & \underline{62.1} & \underline{61.3} & \underline{63.1} & \textbf{82.9} & \underline{80.3} & \underline{81.3} & \underline{68.9} & \underline{50.3} & \underline{53.6} & \underline{55.3} & \underline{55.8} & 48.7 & \underline{52.7} & \underline{63.7} & \underline{58.7}\\
\method{} & \textbf{61.7} & \textbf{64.1} & \textbf{65.0} & \textbf{64.8} & \underline{81.1} & \textbf{83.8} & \textbf{86.1} & \textbf{83.7} & \textbf{52.1} & \textbf{55.3} & \textbf{55.7} & \textbf{57.1} & \textbf{73.7} & \textbf{73.1} & \textbf{77.1} & \textbf{67.6} \\
\toprule
\makecell[c]{\multirow{2}{*}{\textbf{Method}}} & \multicolumn{4}{c}{\textbf{ArXiv}} & \multicolumn{4}{c}{\textbf{DM Mathematics}} & \multicolumn{4}{c}{\textbf{HackerNews}} & \multicolumn{4}{c}{\textbf{Average}} \\
\cmidrule(lr){2-5}\cmidrule(lr){6-9}\cmidrule(lr){10-13}\cmidrule(lr){14-17}
& 160M & 1.4B & 2.8B & 6.9B & 160M & 1.4B & 2.8B & 6.9B & 160M & 1.4B & 2.8B & 6.9B & 160M & 1.4B & 2.8B & 6.9B \\
\midrule
\rowcolor{gray!30}
\multicolumn{17}{c}{\texttt{Gray-Box}}\\
Loss & 74.4 & \textbf{77.6} & \textbf{78.0} & \textbf{78.4} & \underline{94.6} & \underline{93.3} & 92.8 & \underline{92.9} & \underline{57.9} & \underline{59.3} & \textbf{60.5} & \textbf{60.7} & \textbf{72.2} & \textbf{73.5} & \textbf{74.1} & \textbf{74.4}\\
Reference & 57.6 & 71.6 & 71.5 & 71.9 & 61.2 & 48.3 & 44.9 & 44.9 & 51.5 & 53.7 & 57.1 & 57.4 & 58.5 & 62.2 & 62.3 & 61.7\\
Lowercase & 69.2 & 74.0 & 75.5 & 75.8 & 80.7 & 75.9 & 78.2 & 72.6 & 57.7 & 59.0 & 59.9 & \underline{60.2} & 67.3 & 68.8 & 70.0 & 69.6\\
Zlib & \underline{74.5} & \underline{77.2} & 77.7 & \underline{78.0} & 80.9 & 81.0 & 81.2 & 81.0 & 57.5 & 58.4 & 59.3 & 59.4 & 69.6 & 71.2 & 71.8 & 72.1\\
Neighbor & 70.8 & 74.9 & 75.5 & 76.1 & 79.7 & 71.4 & 73.2 & 71.5 & 56.7 & 57.5 & 58.4 & 58.2 & 68.7 & 69.0 & 69.7 & 69.7\\
Min-K\% & 68.6 & 74.2 & 65.3 & 75.6 & 93.1 & \underline{93.3} & \underline{92.9} & \textbf{93.0} & 55.0 & 56.9 & 57.9 & 58.6 &70.0 & 72.5 & 71.8 & \underline{73.7}\\
Min-K\%++ & 52.4 & 64.1 & 65.5 & 65.0 & 76.9 & 76.0 & 70.2 & 73.5 & 53.7 & 56.0 & 57.4 & 58.9 & 60.7 & 65.4 & 66.1 & 67.1\\
ReCaLL & \textbf{76.5} & 76.5 & \underline{77.9} & 76.5 & \textbf{96.7} & \textbf{94.4} & \textbf{94.2} & 91.8 & \textbf{59.1} & \textbf{59.5} & \underline{60.1} & \textbf{60.7} & \underline{71.7} & \underline{72.8} & \underline{73.6} & 73.2\\
PETAL & 68.4 & 70.9 & 71.8 & 72.4 & 86.5 & 86.9 & 86.5 & 86.8 & 57.1 & 58.0 & 58.5 & 58.4 & 66.5 & 67.9 & 68.6 & 68.8\\
\midrule
\rowcolor{gray!30}
\multicolumn{17}{c}{\texttt{Black-Box}}\\
SaMIA & 43.6 & 44.6 & 45.4 & 45.9 & \underline{82.3} & \underline{85.3} & 79.9 & 73.4 & 49.3 & 49.6 & 49.5 & 50.1 & 54.4 & 55.9 & 55.8 & 55.2\\
\method{}$^*$ & \underline{62.9} & \underline{62.5} & \underline{54.5} & \underline{61.3} & 79.0 & 80.5 & \underline{81.4} & \underline{75.9} & \underline{49.8} & \underline{52.4} & \underline{52.8} & \underline{52.2} & \underline{61.6} & \underline{63.4} & \underline{64.3} & \underline{62.3}\\
\method{} & \textbf{72.6} & \textbf{73.8} & \textbf{67.8} & \textbf{73.5} & \textbf{87.6} & \textbf{90.9} & \textbf{90.3} & \textbf{91.3} & \textbf{54.3} & \textbf{54.0} & \textbf{53.0} & \textbf{53.8} & \textbf{69.0} & \textbf{70.7} & \textbf{70.7} & \textbf{70.3} \\
\bottomrule
\end{tabular}
}
\caption{AUC results on MIMIR benchmark. \textbf{Bolded} and \underline{underlined} numbers are the best and second-best results within each column for the given length and setting. $^*$ means that \method{} uses Eq.~\ref{eqn:empirical} rather than Eq.~\ref{eqn:word-score} for scoring.}
\label{tab:mimir}
\vspace{-10pt}
\end{table*}

%% file: tables/wikimia-25.tex
\begin{table*}[t!]
\centering
\resizebox{\linewidth}{!}{
\setlength{\tabcolsep}{5pt}
\begin{tabular}{lcccccccccc}
\toprule
\makecell[c]{\multirow{2}{*}{\textbf{Method}}} &
\multicolumn{2}{c}{\textbf{Pythia-6.9B}} &
\multicolumn{2}{c}{\textbf{Qwen3-8B-Base}} &
\multicolumn{2}{c}{\textbf{Claude-4.5-Haiku}} &
\multicolumn{2}{c}{\textbf{Gemini-2.5-Flash}} &
\multicolumn{2}{c}{\textbf{GPT-5-Chat}} \\
\cmidrule(lr){2-3}\cmidrule(lr){4-5}\cmidrule(lr){6-7}\cmidrule(lr){8-9}\cmidrule(lr){10-11}
& AUC & TPR@5\%FPR & AUC & TPR@5\%FPR & AUC & TPR@5\%FPR & AUC & TPR@5\%FPR & AUC & TPR@5\%FPR \\
\midrule
\multicolumn{11}{>{\columncolor{gray!30}}c}{\texttt{Grey-Box}} \\
Loss & 73.0 & 36.3 & 59.1 & 11.5 & - & - & - & - & - & -\\
Reference & 65.2 & 8.9 & 53.9 & 3.5 & - & - & - & - & - & -\\
Lowercase & 70.7 & 23.0 & 59.4 & 19.5 & - & - & - & - & - & -\\
Zlib & 72.2 & 31.9 & 57.5 & 8.9 & - & - & - & - & - & -\\
Min-K\% & \underline{77.3} & 38.9 & 56.8 & 14.2 & - & - & - & - & - & -\\
Min-K\%++ & 72.7 & \underline{41.6} & 62.7 & \underline{20.4} & - & - & - & - & - & -\\
ReCaLL & \textbf{96.1} & \textbf{78.8} & \textbf{91.0} & \textbf{43.4} & - & - & - & - & - & -\\
PETAL & 72.2 & 30.1 & \underline{62.8} & 18.6 & - & - & \textbf{67.1} & \textbf{23.9} & \textbf{78.0} & \textbf{35.4}\\
\midrule
\multicolumn{11}{>{\columncolor{gray!30}}c}{\texttt{Black-Box}} \\
SaMIA  & 57.8 & \underline{8.8} & 50.3 & 2.7 & 62.6 & 8.0 & 57.0 & 15.0 & 65.7 & 13.3 \\
SimMIA$^*$ & \textbf{90.6} & \textbf{30.1} & \underline{62.9} & \textbf{5.3} & \underline{80.8} & \underline{37.2} & \textbf{77.4} & \textbf{32.7} & \textbf{90.2} & \textbf{71.7}\\
SimMIA & \underline{85.0} & \underline{8.8} & \textbf{74.9} & \underline{4.4}  & \textbf{86.7} & \textbf{51.3} & \underline{75.4} & \underline{29.2} & \underline{75.7} & \underline{21.2}\\
\bottomrule
\end{tabular}
}
\caption{Results on \benchmark{} benchmark. \textbf{Bolded} and \underline{underlined} numbers are the best and second-best results within each column for the given length and setting. $^*$ means that \method{} uses Eq.~\ref{eqn:empirical} rather than Eq.~\ref{eqn:word-score} for scoring.}
\label{tab:wikimia25}
\vspace{-10pt}
\end{table*}

%% file: tables/ablation_study.tex
\begin{table}[t!]
\centering
\small
\resizebox{\linewidth}{!}{
\begin{tabular}{l|ccc}
\toprule
\makecell[c]{\textbf{Method}} & \textbf{Len. 32} & \textbf{Len. 64} & \textbf{Len. 128} \\
\midrule
\method{} & 70.1 & 75.4 & 73.2\\
w/o Word-by-Word Sampling & 55.8 & 50.1 & 47.7\\
w/o Relative Aggregation & 57.4 & 53.7 & 57.0\\
\bottomrule
\end{tabular}
}
\caption{Ablation study on WikiMIA.}
\label{tab:ablation}
\vspace{-10pt}
\end{table}

%% file: figures/num_shots_samples.tex
\begin{figure*}[t!]
\centering
\hspace*{\fill}
{
    \centering
    \input{figures/prefix_ratio.tex}
}\hfill   
{    
    \input{figures/num_samples}
}\hfill
{   
    \centering
    \input{figures/num_shots}
}
\hspace*{\fill}
\caption{WikiMIA (length 32) results under various prefix ratios, sample sizes, and non-member shots.}
\label{fig:combined_numshots}
\vspace{-10pt}
\end{figure*}
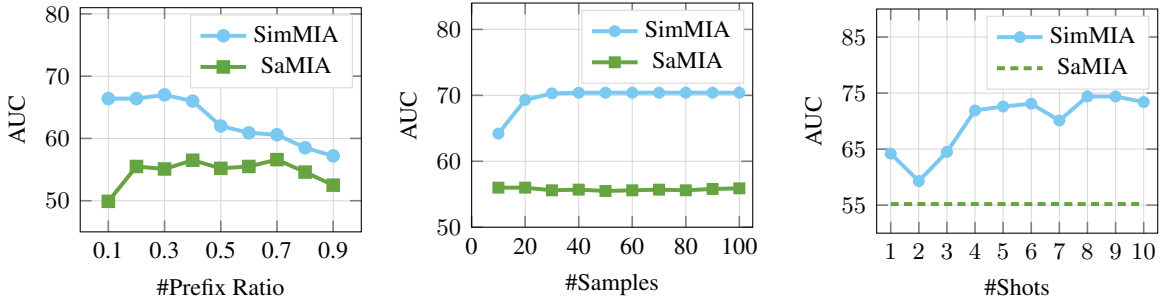

%% file: figures/prefix_ratio.tex
\begin{tikzpicture}
\begin{axis}[
    width=0.33\linewidth,
    xlabel={\#Prefix Ratio},
    ylabel={AUC},
    xmin=0.0, xmax=1.0,
    ymin=45, ymax=81,
    xtick={0.1,0.3,0.5,0.7,0.9},
    xticklabels={0.1,0.3,0.5,0.7,0.9},
    ytick={50,60,70,80},
    ymajorgrids=true,
    xmajorgrids=true,
    grid style={gray!30},
    axis lines=box,
    tick style={semithick},
    xlabel near ticks,
    ylabel near ticks,
    label style={font=\small},
    tick label style={font=\small},
    legend style={at={(0.97,0.97)}, anchor=north east, draw=gray!30, font=\small},
]
\addplot[line width=1.5pt, mark=*, mark size=1.8pt, color=ourblue, mark options={fill=ourblue}]
    coordinates {(0.1,66.4)(0.2,66.4)(0.3,67.0)(0.4,66.0)(0.5,62)(0.6,60.9)(0.7,60.6)(0.8,58.5)(0.9,57.2)};
\addlegendentry{SimMIA}

\addplot[line width=1.5pt, mark=square*, mark size=1.8pt, color=ourgreen, mark options={fill=ourgreen}]
    coordinates {(0.1, 49.9)(0.2,55.5)(0.3,55.1)(0.4,56.5)(0.5,55.2)(0.6,55.5)(0.7,56.6)(0.8,54.6) (0.9,52.5)};
\addlegendentry{SaMIA}

\end{axis}
\end{tikzpicture}
\label{fig:prefix_ratio}

%% file: figures/num_samples.tex
\begin{tikzpicture}
\begin{axis}[
    width=0.33\linewidth,
    xlabel={\#Samples},
    ylabel={AUC},
    xmin=0, xmax=10.5,
    ymin=50, ymax=84,
    xtick={0,2,4,6,8,10},
    xticklabels={0,20,40,60,80,100},
    ytick={50,60,70,80},
    ymajorgrids=true,
    xmajorgrids=true,
    grid style={gray!30},
    axis lines=box,
    tick style={semithick},
    xlabel near ticks,
    ylabel near ticks,
    label style={font=\small},
    tick label style={font=\small},
    legend style={at={(0.97,0.97)}, anchor=north east, draw=gray!30, font=\small},
]
\addplot[line width=1.5pt, color=ourblue, mark=*, mark size=1.5pt, mark options={fill=ourblue}]
    coordinates {(1,64.2)(2,69.3)(3,70.3)(4,70.4)(5,70.4)(6,70.4)(7,70.4)(8,70.4)(9,70.4)(10,70.4)};
\addlegendentry{SimMIA}
\addplot[line width=1.5pt, color=ourgreen, mark=square*, mark size=1.5pt, mark options={fill=ourgreen}]
    coordinates {
    (1,56.0) (2,56.0) (3,55.6) (4,55.7)(5,55.5)(6,55.6)(7,55.7)(8,55.6)(9,55.8)(10,55.9)
    };
\addlegendentry{SaMIA}
\end{axis}
\end{tikzpicture}
\label{fig:numsamples}

%% file: figures/num_shots.tex
\begin{tikzpicture}
\begin{axis}[
    width=0.33\linewidth,
    xlabel={\#Shots},
    ylabel={AUC},
    xmin=0.5, xmax=10.5,
    ymin=50, ymax=90,
    xtick={1,2,3,4,5,6,7,8,9,10},
    ytick={55, 65, 75, 85},
    ymajorgrids=true,
    xmajorgrids=true,
    grid style={gray!30},
    axis lines=box,
    tick style={semithick},
    xlabel near ticks,
    ylabel near ticks,
    label style={font=\small},
    tick label style={font=\small},
    legend style={at={(0.97,0.97)}, anchor=north east, draw=gray!30, font=\small},
]
\addplot[line width=1.5pt, mark=*, mark size=1.5pt, color=ourblue, mark options={fill=ourblue}]
    coordinates {
    (1,64.2) (2,59.3) (3,64.5) (4,71.9) (5,72.6)
    (6,73.1) (7,70.1) (8,74.4) (9,74.4) (10,73.4)
    };
\addlegendentry{SimMIA}
\addplot [
    line width=1.5pt,
    densely dashed,
    color=ourgreen,
    mark=none
] 
    coordinates {
    (1,55.2) (10,55.2)
    };
\addlegendentry{SaMIA}

\end{axis}
\end{tikzpicture}
\label{fig:numshots}

%% file: tables/embedding_models.tex
\begin{table}[t!]
\centering
\small
\resizebox{\linewidth}{!}{
\begin{tabular}{l|ccc}
\toprule
\makecell[c]{\textbf{Embedding Method}} & \textbf{Len. 32} & \textbf{Len. 64} & \textbf{Len. 128} \\
\midrule
Word2Vec & 60.5 & 55.6 & 60.7\\
fastText & 70.6 & 71.9 & 72.8\\
\midrule
bge-large-en-v1.5    & 70.5 & 75.5 & 75.1\\
all-MiniLM-L6-v2     & 70.1 & 75.4 & 73.2 \\
UAE-Large-V1         & 70.2 & 75.2 & 75.2 \\
mxbai-embed-large-v1 & 69.9 & 76.1 & 74.7 \\
\bottomrule
\end{tabular}
}
\caption{WikiMIA results of various embeddings.}
\label{tab:different-embedding-model}
\vspace{-10pt}
\end{table}

%% file: tables/robustness.tex
\begin{table}[t!]
\centering
\small
\resizebox{\linewidth}{!}{
\begin{tabular}{l|ccc}
\toprule
\makecell[c]{\textbf{Method}} & \textbf{Len. 32} & \textbf{Len. 64} & \textbf{Len. 128}\\
\midrule
SaMIA & 55.28$_{0.55}$ & 61.60$_{1.15}$ & 64.26$_{1.69}$\\
\method{} & - & - & - \\
+ Fixed Prefix & 71.26$_{0.89}$ & 74.14$_{0.72}$ & 73.64$_{0.97}$\\
+ Random Prefix & 73.12$_{2.14}$ & 70.08$_{4.29}$ & 75.12$_{2.27}$\\
\bottomrule
\end{tabular}
}
\caption{The mean and standard deviation (in subscript) of five runs in WikiMIA.}
\label{tab:robustness}
\vspace{-10pt}
\end{table}

%% file: tables/prefix_set.tex
\begin{table}[t!]
\centering
\small
\setlength{\tabcolsep}{6pt}
\renewcommand{\arraystretch}{1.1}
\begin{tabular}{l|ccc}
\toprule
\makecell[c]{\textbf{Setting}} & \textbf{Len. 32} & \textbf{Len. 64} &  \textbf{Len. 128}\\
\midrule
Most & 77.7 & 77.9 & 71.3 \\
Moderate & 71.1 & 70.6 & 64.5\\
Least & 58.3 & 60.5 & 56.1\\
Random & 69.8 & 73.4 & 70.1\\ 
\midrule
Fixed & 70.1 & 75.4 & 73.2\\
\bottomrule
\end{tabular}
\caption{Results with prefixes of different similarities.
}
\label{tab:prefix-selection}
\vspace{-10pt}
\end{table}

%% file: tables/expose.tex
\begin{table}[t!]
\centering
\small
\resizebox{\linewidth}{!}{
\begin{tabular}{l|cc}
\toprule
\makecell[c]{\textbf{Metadata}}&\textbf{GPT-5-Chat} &\textbf{GPT-4.1-mini} \\

\midrule
SimMIA (Text Only) & 82.0 & 85.3 \\
+ Tokenization & 83.1 & 87.0 \\
+ Token Log Probs & - & 93.7 \\
\bottomrule
\end{tabular}
}
\caption{Progressive disclosure results on \benchmark{}.}
\label{tab:expose}
\vspace{-10pt}
\end{table}

%% file: tables/close-source-model-versions.tex
\begin{table}[!b]
\centering
\small
\setlength{\tabcolsep}{6pt}
\renewcommand{\arraystretch}{1.1}
\begin{tabular}{l|l}
\toprule
\makecell[c]{\textbf{Model Name}} & \makecell[c]{\textbf{Version}} \\
\midrule
Claude-4.5-Haiku & \texttt{claude-haiku-4-5-20251001} \\
Gemini-2.5-Flash & \texttt{gemini-2.5-flash} \\
GPT-5-Chat & \texttt{gpt-5-chat-latest} \\
GPT-4.1-mini & \texttt{gpt-4.1-mini-2025-04-14} \\
\bottomrule
\end{tabular}
\caption{The versions of proprietary LLMs used.}
\label{tab:model_versions}
\end{table}

%% file: tables/prefix_domain.tex
\begin{table}[t!]
\centering
\small
\begin{tabular}{l|ccc}
\toprule
\makecell[c]{\textbf{Domain}} & \textbf{Len. 32} & \textbf{Len. 64} & \textbf{Len. 128}\\
\midrule
Original         & 70.1 & 75.4 & 73.2 \\
Wikipedia        & 51.3 & 49.7 & 52.2 \\
arXiv            & 57.6 & 52.6 & 59.7 \\
GitHub           & 56.0 & 53.8 & 51.3 \\
\bottomrule
\end{tabular}
\caption{Results of various non-member prefix domains. We sample non-members from the corresponding domains in MIMIR for testing. We truncate each non-member to 32, 64, or 128 words to ensure consistency with the prefix lengths used in WikiMIA.}
\label{tab:different-prefix-domain}
\end{table}